# Enhancing Road Crack Detection Accuracy with BsS-YOLO: Optimizing Feature Fusion and Attention Mechanisms

Jiaze Tang[a], Angzehua Feng[a,*], Vladimir Korkhov[a] and Yuxi Pu[a]

[a]*Department of Applied Mathematics and Control Processes, Saint Petersburg State University, Saint Petersburg 199034, Russia*

ABSTRACT

Effective road crack detection is crucial for road safety, infrastructure preservation, and extending road lifespan, offering significant economic benefits. However, existing methods struggle with varied target scales, complex backgrounds, and low adaptability to different environments. This paper presents the BsS-YOLO model, which optimizes multi-scale feature fusion through an enhanced Path Aggregation Network (PAN) and Bidirectional Feature Pyramid Network (BiFPN). The incorporation of weighted feature fusion improves feature representation, boosting detection accuracy and robustness. Furthermore, a Simple and Effective Attention Mechanism (SimAM) within the backbone enhances precision via spatial and channel-wise attention. The detection layer integrates a Shuffle Attention mechanism, which rearranges and mixes features across channels, refining key representations and further improving accuracy. Experimental results show that BsS-YOLO achieves a 2.8% increase in mean average precision (mAP) for road crack detection, supporting its applicability in diverse scenarios, including urban road maintenance and highway inspections.

## 1. Introduction

Detecting road cracks is a global priority crucial for maintaining road safety, generating economic benefits, and maintaining long-term infrastructure stability. Timely detection and repair of road cracks effectively prevent traffic accidents, safeguarding lives and property Shi, Cui, Qi, Meng & Chen (2016); Guo & Zhang (2022) These cracks often signal the early stages of aging in roads, bridges, tunnels, and other infrastructures Liu, Zhu, Xia, Zhao & Long (2022a); Sekar & Perumal (2022).According to the World Health Organization, road traffic accidents cause significant global economic losses each year, accounting for up to 3% of the GDP in many countries. Around 1.3 million individuals lose their lives each year due to road traffic accidents, with another 20 to 50 million suffering non-fatal injuries Lancet (2022); WHO (2023). The aging infrastructure, excessive pressure from overloaded vehicles, environmental factors, and design flaws accelerate crack formation, posing risks to driving safety and impacting the economy and society Feng, Xiao, Li, Pei, Sun, Ma, Shen & Ju (2020). Therefore, promptly detecting road cracks is a critical and pressing issue.

Road crack detection techniques can be classified into two main categories: traditional methods and approaches based on computer vision. Traditional methods include manual detection and sensor-based techniques. Early methods employed sensors such as cameras and LiDAR to monitor and scan road surfaces in real-time. Additionally, road maintenance personnel periodically inspect roads, using visual observation and manual recording to detect cracks Hu, Hu, Yang, Huang & Li (2021); Ashraf, Sophian, Shafie, Gunawan & Ismail (2023). Researchers also utilized road profilometers, scanning road surfaces with lasers or ultrasonic sensors mounted on vehicles to detect and analyze cracks Dong, Chen, Dong & Ni (2021). However, these traditional methods have several drawbacks. Sensor-based methods are costly, involve complex data processing, and are significantly affected by environmental factors. Even though manual detection is uncomplicated, it is typically time-consuming, requires considerable manpower, and is subject to personal bias Quan, Sun, Zhang & Zhang (2019); Li, Liu, Zhao, Qiao & Ren (2020). Consequently, computer vision methods capable of timely and intelligent road crack detection have gained popularity.

Computer vision approaches are primarily split into traditional machine learning and deep learning techniques Cong, Lei, Fu, Cheng, Lin & Huang (2018).Traditional machine learning methods include Support Vector Machines

*Corresponding author
✉ st122421@student.spbu.ru (J. Tang); st122387@student.spbu.ru (A. Feng); v.korkhov@spbu.ru (V. Korkhov); st122375@student.spbu.ru (Y. Pu)
ORCID(s): 0009-0002-5594-3244 (J. Tang); 0009-0002-7672-1800 (A. Feng); 0000-0003-2458-3194 (V. Korkhov); 0009-0004-2807-9635 (Y. Pu)

(SVM), Random Forests, and k-Nearest Neighbors (k-NN) algorithms Salazar-Rojas, Cejudo-Ruiz & Calvo-Brenes (2022); Sun, Cheng, Zhang, Mohan, Ye & De Schutter (2023); Zhang, Cao, Wang & Li (2019). For example, Yu, Rashidi, Samali, Yousefi & Wang (2021) employed SVM and D-S fusion algorithms to identify cracks, achieving an accuracy of 96.25%. Peng, Yang, Zheng, Zhang, Wang, Yan, Wang & Li (2020) used a three-threshold method of random structured forests to enhance image information blocks through channel feature pairwise differences, achieving an accuracy of 95.95%. Müller, Karathanasopoulos, Roth & Mohr (2021) proposed an image-based machine learning method to classify samples with and without cracks, reaching 99% accuracy. Lei & Zuo (2009) utilized the k-NN classification algorithm to identify crack faults in gears, successfully identifying different crack severities and grading the faults. However, traditional machine learning methods have limitations, including a high dependence on manual feature engineering, difficulty in automatically extracting complex features, sensitivity to data quality and quantity, and challenges in handling high-dimensional data Vijayan, Joy & Shailesh (2021). These limitations restrict the performance of traditional machine learning models, making deep learning methods, which automatically learn features and adapt to large-scale data, more suitable for road crack detection He, Tang, Deng, Zhou, Wang & Li (2023).

Deep learning techniques are classified into two-stage and one-stage methodologies Liu, Ouyang, Wang, Fieguth, Chen, Liu & Pietikäinen (2020). Two-stage methods initially produce candidate regions that are probable to contain the target, followed by classifying and precisely determining the boundaries of the target. Typical two-stage object detection methods include the R-CNN series, such as R-CNN, Fast R-CNN, Faster R-CNN, and Mask R-CNN. For instance, Li, Yu, Li, Yang, Wang & Peng (2023b) employed Faster R-CNN along with Unmanned Aerial Vehicles (UAVs) in crack detection experiments, which resulted in an accuracy of 92.03%, a recall rate of 96.26%, and an F1 score of 94.10%. Huyan, Li, Tighe, Zhai, Xu & Chen (2019) proposed CrackDN based on the Fast R-CNN architecture for detecting sealed and unsealed cracks in complex road backgrounds, achieving an average precision exceeding 0.90, outperforming SSD300. Du, Lu & Li (2023) used Mask R-CNN to automatically detect cracks from electric image logs, demonstrating an accuracy of 96% and a recall of 92%. However, two-stage deep learning methods require high computational power, have low efficiency and real-time performance, and may misdetect small targets, limiting their applicability in some real-world scenarios. As a result, the detection of road fractures has become dependent on one-stage deep learning approaches.

One-stage deep learning methods include SSD (Single Shot MultiBox Detector), the YOLO (You Only Look Once) series, and RetinaNet. Sun, Ai, Wang & Zhang (2021) employed an improved SSD algorithm for small target detection, achieving better results than the baseline SSD on the Tsinghua-Tencent 100K and Caltech Pedestrian datasets. However, weak segmentation masks limited the segmentation branch's precision and handling of complex backgrounds, and further optimization for small target detection and scale fusion was not considered. Li, Ren, Wang, Wang, Wang & Du (2023a) utilized YOLOv3 for scale fusion, performing well in detecting small and overlapping coal gangue targets with an average precision of 98.97%. However, YOLOv3's accuracy issues still require improvement in complex environments. YOLOv5 introduced several architectural improvements, including enhanced Feature Pyramid Networks (FPN) and Path Aggregation Networks (PAN), improving multi-scale feature fusion. Zhou, Su, Li & Dai (2023) developed a drone target detection algorithm based on YOLOv5, achieving better speed and accuracy than YOLOv3, with superior accuracy and precision in complex environments.

Despite these advances, YOLOv5 employs relatively basic strategies during training, with less advanced data augmentation and regularization techniques than YOLOv7, impacting the model's generalization ability and stability. YOLOv7 introduced an optimized backbone network, improving feature extraction capabilities, particularly in handling complex backgrounds and small targets Ye, Qu, Tao, Dai, Mao & Jin (2023). Zhao, Shao, Liu, Yang, Zhang & Zhang (2024) used a YOLOv7-based model for detecting small objects in drone images, achieving impressive mAP50 scores of 56.8% and 94.6% on the VisDrone2019 and NWPU VHR-10 datasets, respectively. YOLOv8 offers significant improvements in speed and accuracy. Liu, Li, Wang & Zhu (2024) proposed a YOLOv8-based algorithm to address various types of damage, including cracks and wear. The thin-neck structure better fused multi-scale features with background information, resulting in a mean average precision of 79.9%. However, YOLOv8 may still struggle with small target detection compared to some specially optimized models, and its deeper network and larger parameter set reduce training efficiency.

Previous studies have made significant progress in road crack detection, but challenges remain. For instance, current methods still struggle with varying target scales and complex backgrounds. To address these issues, this paper proposes the BsS-YOLOv8 model. Specifically, BiFPN enhances multi-scale feature expression through efficient cross-scale feature fusion, improving detection accuracy across different target scales. Additionally, SimAM integrates spatial adaptive modules and channel attention mechanisms, reducing computational complexity and enhancing model precision. Finally, Shuffle Attention strengthens the model's ability to record local and global data improves the detection accuracy in complex backgrounds. Compared to existing methods, BsS-YOLOv8 significantly improves detection accuracy, making it appropriate for various practical applications, such as urban road maintenance and highway inspections.

## 2. Theories and Methods
### 2.1. YOLOv8

YOLOv8 is an advanced object detection model. It performs outstandingly in object detection, instance segmentation, and image classification. This version introduces several enhancements aimed at improving both accuracy and speed. YOLOv8 includes four model variants: YOLOv8n, YOLOv8m, YOLOv8l, and YOLOv8x.The structure diagram of YOLOv8n is shown in Figure 1. Each model is constituted by three major components, namely Backbone, Neck, and Head.

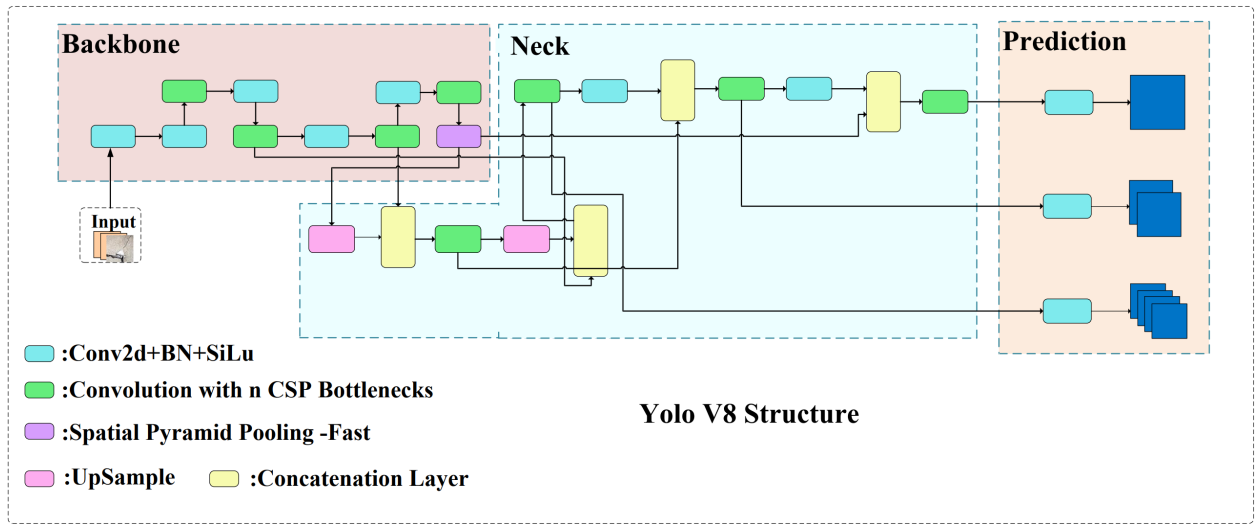

**Figure 1:** The Structure of YOLOv8n

The main function of the Backbone is to receive the input image and extract its features. This typically involves convolutional layers for feature extraction from low to high levels, batch normalization layers to speed up convergence and stabilize the training process, and activation functions (such as ReLU or SiLU) to introduce non-linearity and enhance the model's expressiveness.

The Neck is responsible for fusing features at different scales, multi-scale feature maps are generated for the detection head. It generally includes a FPN, which uses lateral connections and upsampling to create pyramid feature maps, enabling the detection of objects at various scales. Additionally, a PAN is incorporated to enhance feature fusion and propagation through multiple upsampling and downsampling operations, thereby improving detection performance.

The Head processes the multi-scale feature maps to yield the final detection results, encompassing bounding boxes, class probabilities, and other pertinent information. It typically consists of convolutional, classification, and regression layers. Specific implementations may use Anchor or Anchor-Free mechanisms for target localization. The Anchor mechanism enhances detection accuracy by matching predefined anchor boxes with real targets, while the Anchor-Free mechanism directly predicts the center points and sizes of targets, simplifying the model by eliminating the need for predefined anchor boxes.

### 2.2. Bidirectional Feature Pyramid Network (BiFPN)

In road crack detection, variations in crack size and shape present significant challenges. YOLOv8 uses a strategy that integrates multi-scale feature information through FPN and PAN. This method skillfully combines deep high-level semantic features with shallow high-resolution features, capturing the target information at several scales. However, this strategy leads to an increase in the number of parameters and imposes higher computational demands. Additionally, the FPN and PAN structures may contribute to overfitting, reduce generalization ability, and fail to fully consider the weight differences between input features.

To overcome these challenges, this study introduces an optimized PAN-based BiFPN Tan, Pang & Le (2020), as depicted in Figure 2. This network effectively merges features at different scales through its bidirectional structure, significantly enhancing detection capabilities for small objects and improving overall model accuracy Cao, Fu, Zhu & Cai (2022). Furthermore, by optimizing feature flow and reducing computational complexity, BiFPN enhances the model's operational efficiency, achieving superior performance in scenarios that require rapid response.

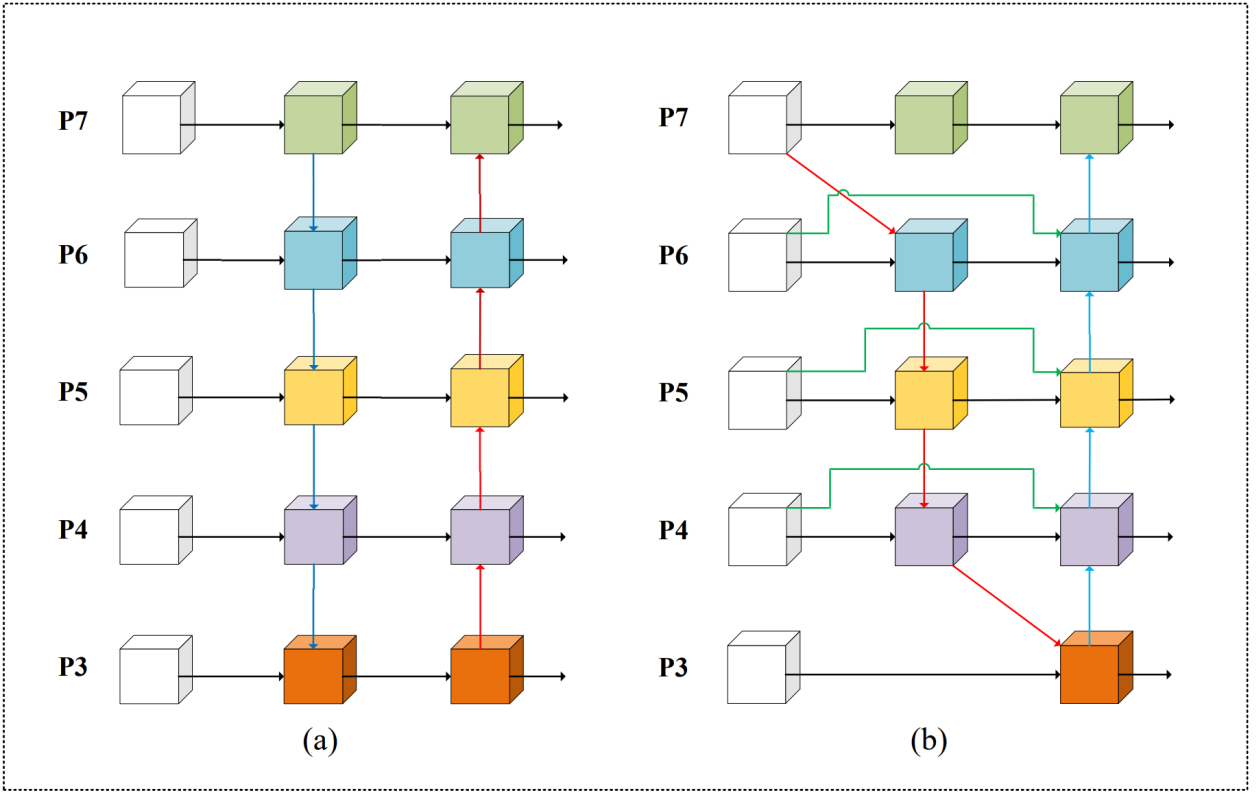

**Figure 2:** Different neck architecture designs (a) PANet (b) BiFPN.

The optimization process first focused on the PAN structure, where it was observed that some nodes received unidirectional input without contributing significantly to deep feature fusion. These low-contribution nodes increased computational and parameter burdens while adding little to overall feature fusion. To improve detection efficiency, a strategy was adopted to remove these nodes. Additionally, the bidirectional information flow among feature layers was improved by adding skip connections between input and output nodes at the same level, resulting in more effective information fusion, as illustrated in Figure 3 of the BiFPN structure. This adjustment improves the comprehensive utilization of information while maintaining manageable computational costs.

To improve the precision of and classification at diverse resolutions, this study introduces a fast-weighted normalization fusion technique. This method assigns additional weights to input features, normalizing each weight by their sum to achieve effective feature fusion. This method significantly improves the performance of road crack detection tasks by enabling the network to recognize the relative importance of each feature layer. The weighted strategy ensures that all weight values fall between 0 and 1, as outlined in Formula (1).

$$O = \sum_i \frac{w_i}{\varepsilon + \sum_j w_j} I_i \tag{1}$$

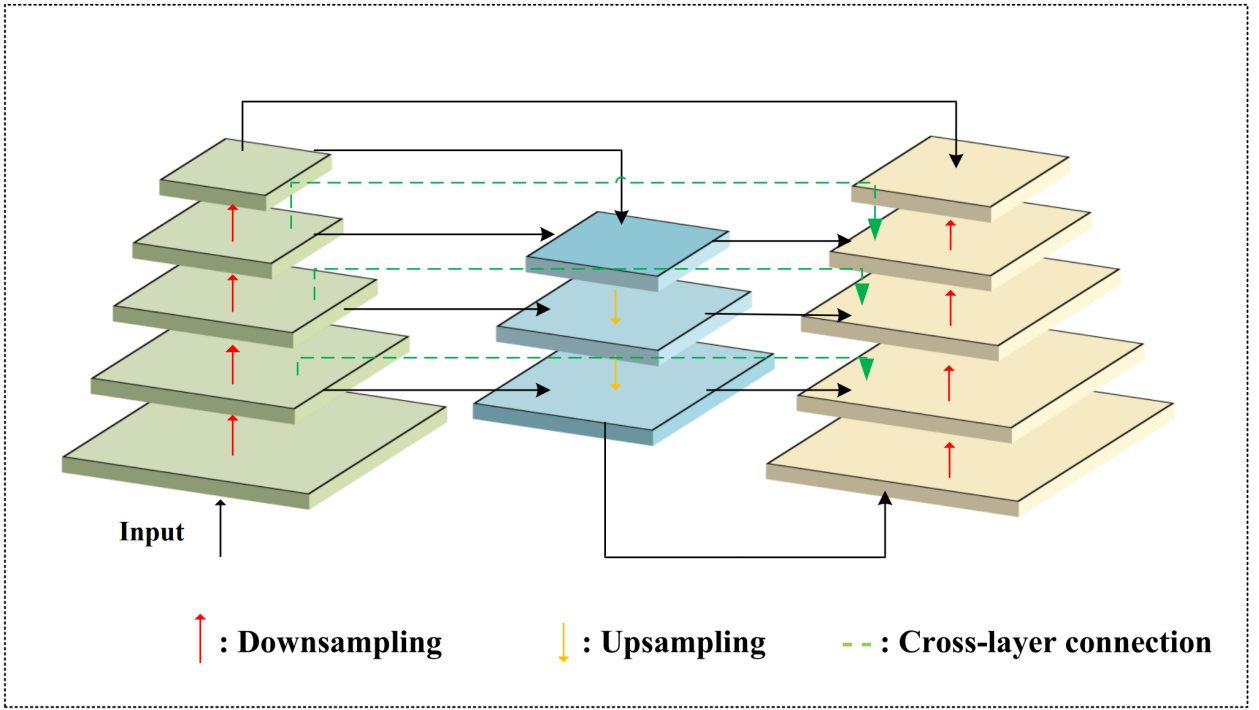

**Figure 3:** The Structure of BiFPN

In this context, $O$ represents the output features, $I_i$ represents the input features, and $w$ denotes the node weights. The learning rate, $\varepsilon$ is set to 0.0001 to prevent unstable results.

### 2.3. Simple and Effective Attention Mechanism (SimAM)

In practical road crack detection, complex backgrounds often demand higher accuracy from crack detection models. The YOLOv8 model's enhanced detection performance has led to its widespread application in a variety of contexts. However, traditional attention mechanisms, while effective at improving detection, typically require extensive hyperparameter tuning, which increases computational complexity and the number of parameters. Additionally, these mechanisms rely on complex computations, which demand high computational resources.

To tackle these challenges, this study integrates the SimAM into the YOLOv8 design. SimAM simplifies the attention mechanism, enabling effective feature extraction and information fusion while reducing the need for hyperparameter tuning and computational resources Liu, Quijano & Crawford (2022b). As illustrated in Figure 4, the attention mechanism is specifically integrated into the backbone of YOLOv8, enhancing detection accuracy without increasing model complexity.

The YOLOv8 model, enhanced with SimAM, retains its lightweight nature while improving both information extraction efficiency and detection performance. Consequently, the SimAM-equipped YOLOv8 delivers more accurate detection results in complex environments, significantly reducing computational costs while maintaining performance. This makes the model suitable for a broader range of practical applications, including urban road maintenance and highway inspection.

SimAM calculates an energy function for every neuron to estimate its significance, enabling the network to learn more discriminative neurons and better extract features without adding extra parameters. This approach highlights important features and suppresses irrelevant information, enabling the model to better learn discriminative feature representations. SimAM is grounded in neuroscience theory and uses an energy function to extract key features Yang, Zhang, Li & Xie (2021). The energy function of each neuron is described in Equation (2).

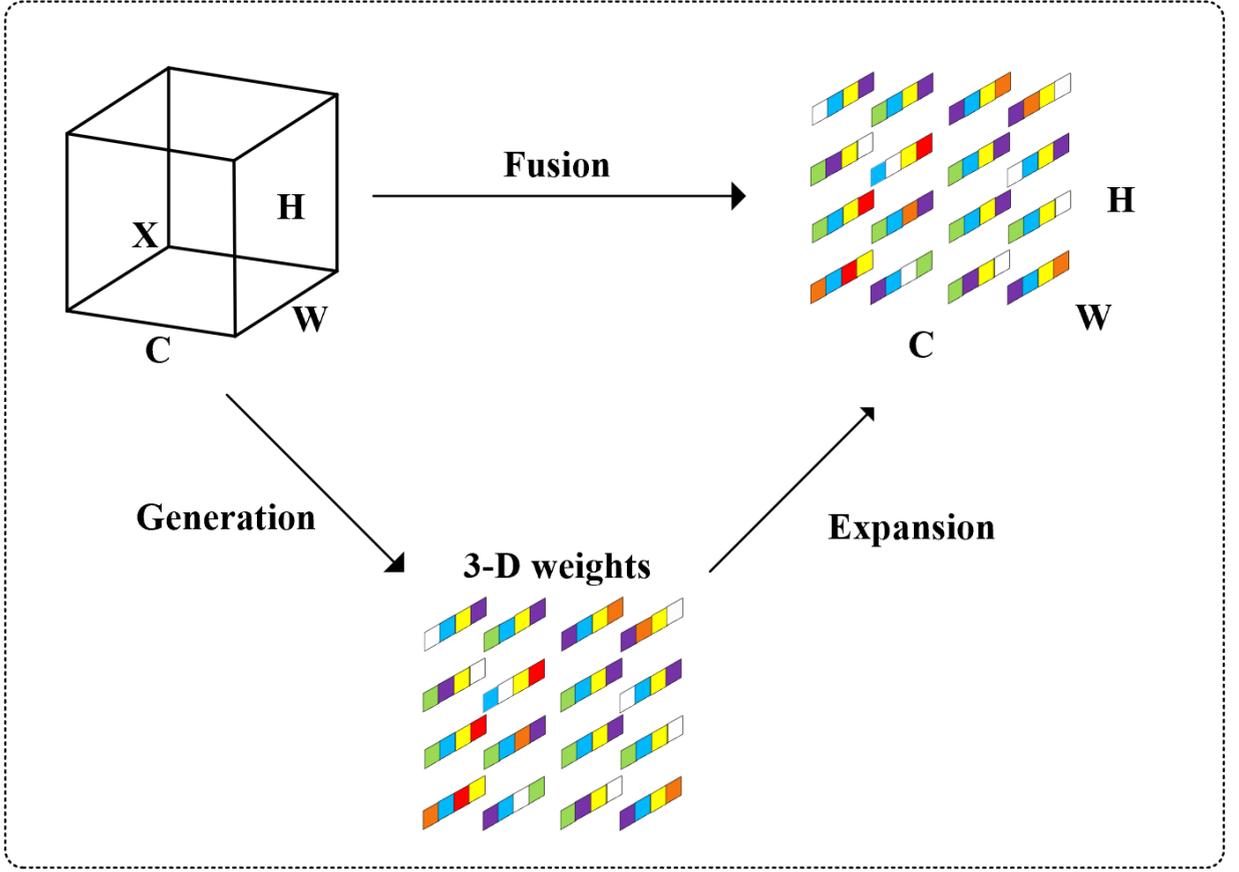

**Figure 4**: SimAM with full 3D weights for attention

$$e_t(\omega_t, b_t, \vec{y}, x_i) = (y_t - \hat{t})^2 + \frac{1}{M-1}\sum_{i=1}^{M-1}(y_0 - \hat{x}_i)^2 \tag{2}$$

In this context, $t$ and $x_i$ represent the target neuron and other neurons within a single channel of the input feature $X \in \mathbb{R}^{C \times H \times W}$, where the index $i$ refers to the spatial dimension. The linear transformations of t and $x_i$ are denoted by $\hat{t} = \omega_t + b_t$ and $\widehat{x_i} = \omega_t x_i + b_t$, with $\omega_t$ and $b_t$ representing the weights and biases of the transformation, respectively. The number of neurons in the channel is $M = H \times W$. By minimizing this equation, Equation (1) essentially it determines the linear separability of the target neuron $t$ from the other neurons within the same channel. For simplicity, binary labels (1 and -1) are assigned to $y_t$ and $y_0$, and a regularizer is added. The final energy function is provided in Equation (3).

$$e_t(\omega_t, b_t, \vec{y}, x_i) = \frac{1}{M-1}\sum_{i=1}^{M-1}\left(-1 - (\omega_t x_i + b_t)\right)^2 + \left(1 - (\omega_t x_i + b_t)\right)^2 + \lambda \omega_t^2 \tag{3}$$

Theoretically, each channel contains $M = H \times W$ energy functions. Equation (3) has an analytical solution for $\omega_t$ and $b_t$, which can be obtained using Equation (4).

$$\omega_t = -\frac{2(t - \mu_t)}{(t - \mu_t)^2 + \frac{2}{M-1}\sum_{i=1}^{M-1}(x_i - \mu_t)^2 + 2\lambda}$$

$$b_t = -\frac{1}{2}(t + \mu_t)\omega_t \tag{4}$$

Here, $\mu_t = \frac{1}{M-1}\sum_{i=1}^{M-1} x_i$ and $\sigma_t^2 = \frac{1}{M-1}\sum_{i}^{M-1} (x_i - \mu_t)^2$ represent the mean and variance calculated for all neurons within the channel, excluding $t$. Therefore, the minimum energy is expressed as shown in Equation (5).

$$e_t^* = \frac{4(\hat{\sigma}^2 + \lambda)}{(t - \hat{\mu})^2 + 2\hat{\sigma}^2 + 2\lambda} \tag{5}$$

Where $\hat{\mu} = \frac{1}{M}\sum_{i=1}^{M} x_i$ and $\hat{\sigma}^2 = \frac{1}{M}\sum_{i=1}^{M} (x_i - \hat{\mu}_t)^2$. Assuming all pixels have the same distribution, a lower energy of neuron $t$ indicates that it is more distinct from surrounding neurons, thereby increasing its importance. Based on Equation (5), the weight of each neuron is given by $\frac{1}{e_t^*}$.

Therefore, the SimAM attention module can be described by Equation (6).

$$\mathbf{X} = \text{sigmoid}\left(\frac{1}{\mathbf{E}}\right) \odot \mathbf{X} \tag{6}$$

$\mathbf{E}$ aggregates all $\frac{1}{e_t^*}$ values across the channel and spatial dimensions. A sigmoid function is applied to constrain the magnitude of $\mathbf{E}$.

### 2.4. Shuffle Attention (SA)

Attention mechanisms have been widely applied in convolutional neural networks, significantly enhancing performance across various tasks. In vision tasks, attention mechanisms are mainly classified into two categories: channel attention and spatial attention. As illustrated in Figure 5, channel attention selects critical features along the channel dimension of the feature map, enhancing the model's focus on specific features, thereby improving feature extraction efficiency and accuracy. However, calculating attention weights for each channel may increase computational complexity and potentially overlook local spatial information. Spatial attention, in contrast, selects important regions along the spatial dimension of the feature map, capturing spatial dependencies and local features but may neglect global features and also face high computational complexity.

The SA-Net integrates the merits of both channel attention and spatial attention, simultaneously capturing global and local information to improve model performance through adaptive weight assignment. As shown in Figure 6, SA-Net significantly enhances model accuracy and generalization while maintaining a lightweight structure, reducing parameter overhead, It holds extensive application prospects in fields like computer vision and image processing. It can be used for things like classification of images, and image segmentation. By optimizing the attention mechanism, SA-Net groups the input feature map by channels and applies Shuffle Units to implement both spatial and channel attention for each group.

The entire SA module is divided into three steps: Feature Grouping, Mixed Attention, and Feature Aggregation.

#### 2.4.1. Feature Grouping

Feature Grouping divides the input feature $X$ along the channel dimension into $K$ groups, i.e., $X = [X_1, \ldots, X_K]$, where each group is a sub-feature ($X_k \in \mathbb{R}^{C/K \times H \times W}$). During training, each sub-feature $X_i$ progressively captures specific semantic responses. Subsequently, the attention module produces corresponding importance coefficients for each sub-feature.

#### 2.4.2. Channel Attention and Spatial Attention

In the proposed approach, the sub-feature $X_k$ is divided into two branches along the channel dimension, denoted as $X_{i1}$, $X_{i2} \in \mathbb{R}^{C/2K \times H \times W}$, as illustrated in the Split section of Figure 7. The two branches are highlighted in purple and green, respectively. The purple branch implements channel attention to capture inter-channel dependencies, while the green branch captures spatial dependencies between features, generating a spatial attention map. This dual attention mechanism enables the model to concurrently pay attention to both semantic and positional information.

In the channel attention branch, a lightweight design is prioritized, opting for the simplest GAP+Scale+Sigmoid single-layer transformation to minimize parameter count (as SE has more parameters) Zhang & Yang (2021). This transformation is represented in Equation (7).

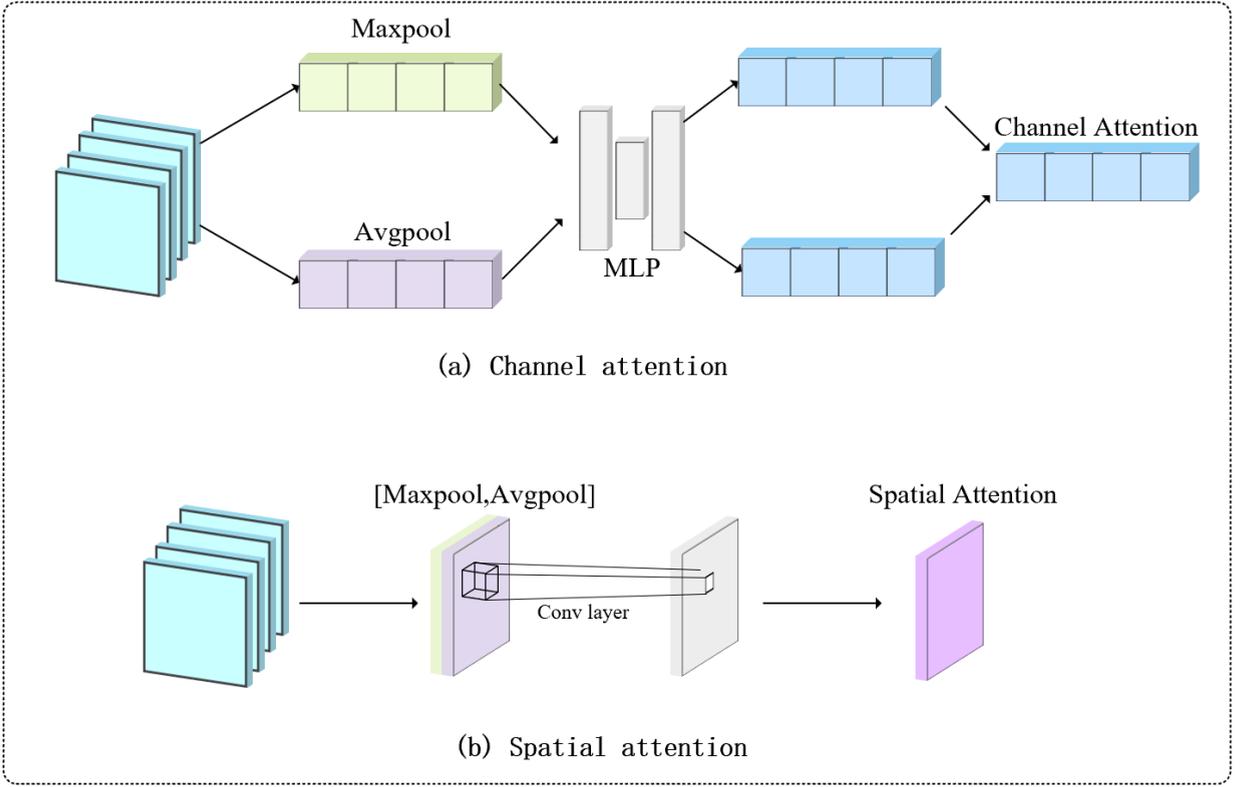

**Figure 5**: Two different models of attention mechanisms .(a) Channel attention (b) Spatial attention

$$g = \Psi_{kp}(X_{k1}) = \frac{1}{H \times W} \sum_{i=1}^{H} \sum_{j=1}^{W} X_{k1}(i,j)$$

$$X'_{k1} = \xi(\Psi_c(g)) \cdot X_{k1} = \xi W_1 g + b_1 \cdot X_{k1} \tag{7}$$

Equation (7) in the channel attention branch involves only two transformation parameters: $W_1 \in \mathbb{R}^{C/2K \times 1 \times 1}$ and $b_1 \in \mathbb{R}^{C/2K \times 1 \times 1}$.

In the spatial attention branch, Group Normalization is first applied to the input feature map, followed by a transformation $\mathcal{F}_c(\bullet)$ to enhance the input representation, as specified in Equation (8).

$$X'_{k2} = \xi(W_2 \cdot GN(X_{k2})) + b_2 \cdot X_{k2} \tag{8}$$

The outputs from both attention mechanisms are then combined as $X'_k = [X'_{k1}, X'_{k2}] \in \mathbb{R}^{C/K \times H \times W}$, with the number of channels matching that of the input.

### 2.4.3. Feature Aggregation

Finally, all sub-features are aggregated. In order to enable information flow between groups along the channel dimension, a "channel shuffle" operation is used. The SA module's final output keeps the same dimensions as the input X. with the channel shuffle operation, as implemented in ShuffleNetv2, ensuring effective interaction between different sub-feature groups. This makes the SA module highly compatible with modern architectures.

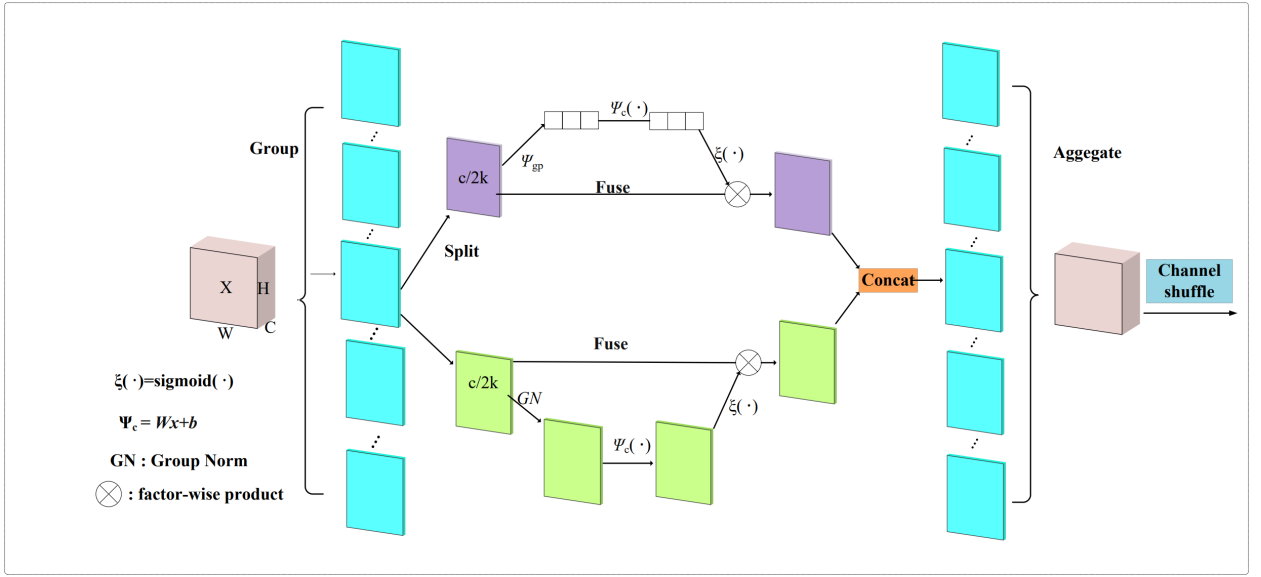

**Figure 6**: The Structure of SA-Net

## 3. Experiment and Analysis
### 3.1. Dataset
The RDD2022 dataset Arya, Maeda, Ghosh, Toshniwal & Sekimoto (2022) is a large-scale road damage detection resource, capturing images from various perspectives, including vehicle-mounted, drone, and handheld devices. It consists of 3900 images, each labeled with specific road damage types such as D00 (longitudinal cracks), D10 (transverse cracks), D20 (alligator cracks), and D40 (potholes). These images capture a variety of road surfaces including cement and asphalt roads. They also display various environmental conditions and different lighting variations from dawn to dusk. To ensure consistency during model training, all images were standardized to a resolution of 640×640 pixels. The dataset is divided into training, validation, and test sets in a ratio of 7:2:1, providing a balanced distribution across the different road damage categories for effective model development and evaluation.

The quantity of these four types of bounding boxes annotating different damage types and the size distribution of annotation boxes are shown in Figure 7. Among them, D00 accounts for the largest number of tags, while D40 has the fewest tags, reflecting a potential imbalance in the dataset. This distribution highlights the need for models to handle such variations effectively during detection tasks.

To fulfill the requirements of the experiment, We used a labeling tool to label four common types of road cracks. The label data of the training set comprises category information, central coordinates (x, y) of the bounding box, and width and height values. All these details are presented in figure 8.

### 3.2. Evaluation Metrics
In order to perform a thorough assessment of the effectiveness of the proposed model for road crack detection, four fundamental evaluation metrics are utilized: Precision (P), Recall (R), Mean Average Precision (mAP), and F1 score (F1). The formulas corresponding to these metrics are presented in Equations (9-12).

$$P = \frac{TP}{TP + FP} \quad (9)$$

$$R = \frac{TP}{TP + FN} \quad (10)$$

$$F1 = 2 \cdot \frac{P \cdot R}{P + R} \quad (11)$$

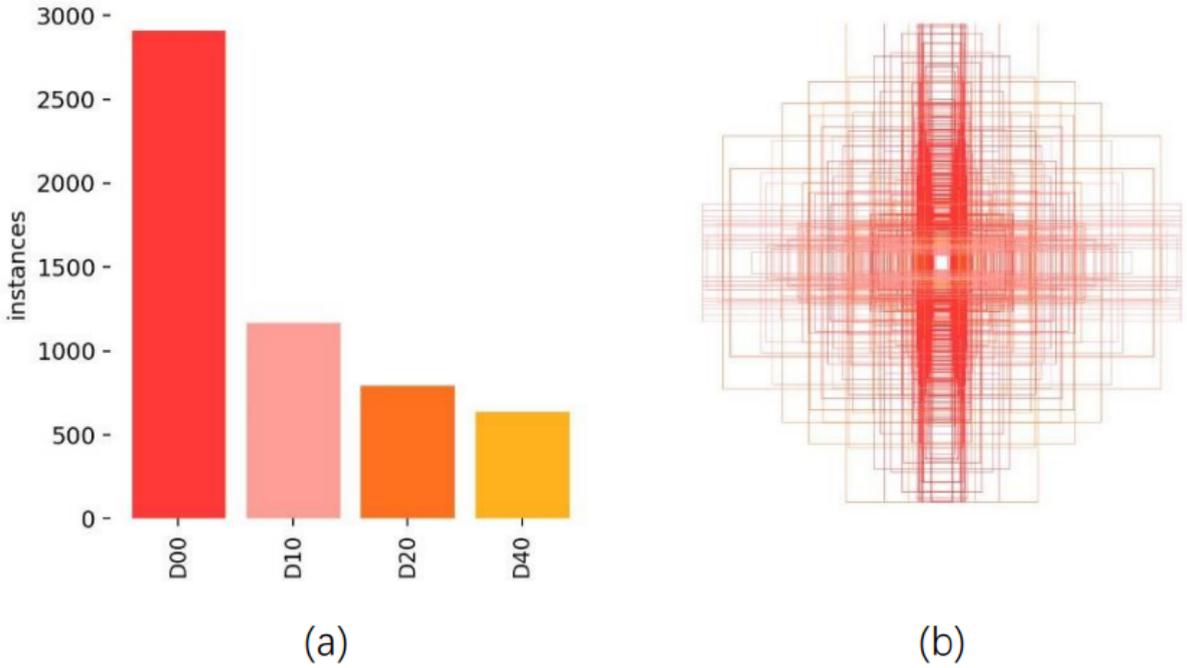

**Figure 7:** Data volume by category and bounding box sizes/quantities. (a) the distribution of training data volume; (b) the size distribution of the labeled boxes.

$$mAP = \frac{1}{Q} \sum_{q=1}^{Q} AP(q) \tag{12}$$

- TP (True Positives): This indicates the quantity of cracks that are accurately detected by the model.
- FP (False Positives): This represents the number of instances that are mistakenly identified as cracks by the model.
- FN (False Negatives): The actual cracks that the model failed to detect.

The mAP is computed as the average of the Average Precision (AP) values for all categories. Here, Q denotes the total number of categories.

### 3.3. Before and After Improvement

To verify the effectiveness of the proposed BsS-YOLO model, we conducted a detailed comparison with the YOLOv8 model. The table presents the performance comparison across four major metrics: Precision, Recall, F1 score, and mAP. From these data, it can be concluded that the results of the BsS-YOLO model have been significantly improved, as shown in Table 1:

**Table 1**
Before and After

| Model | Precision (%) | Recall (%) | F1 (%) | mAP (%) |
|---|---|---|---|---|
| YOLOv8n | 52.1 | 44.3 | 47.9 | 41.5 |
| BsB-YOLO | 57.3 | 47.3 | 51.8 | 44.3 |

The precision of the BsS-YOLO model increased from 52.1% in YOLOv8n to 57.3%, an increase of 5.2 percentage points. This improvement is mainly due to the application of BiFPN, it enhances the model's capability of detecting small targets via bidirectional feature fusion, effectively reducing the false positive rate.

Additionally, the recall rate of the improved modelis increased to 47.3%, it is 3 percentage points higher than the baseline model. This indicates that BsS-YOLO is more comprehensive in detecting cracks, effectively reducing missed detections. The SimAM module was instrumental in this improvement, guiding the model to focus specifically on information crucial to the task, thereby enhancing the model's sensitivity and extraction ability for key information.

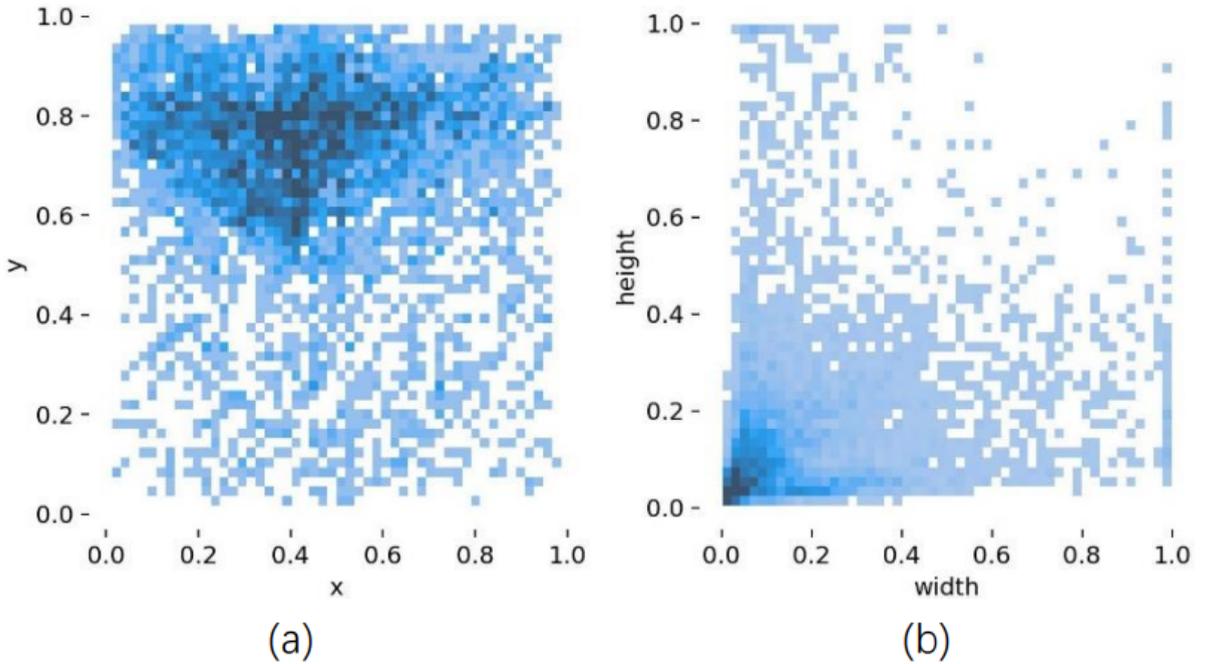

**Figure 8:** The position of the center point of the detection box and the target distribution: (a) center point position relative to the whole image; (b) distribution of the labeled boxes' width and height.

The F1 score for BsS-YOLO improved from 47.9% in the baseline model to 51.8%, an increase of 3.9 percentage points. This reflects the simultaneous improvement in precision and recall, indicating a better balance between these two critical metrics.

The mAP of BsS-YOLO was 44.3%, which was an increase of 2.8 percentage points compared with the 41.5% of YOLOv8n. This improvement demonstrates significantly enhanced overall performance in multi-class crack detection tasks. Through the fusion of channel attention and spatial attention mechanisms, the SA-Net module efficiently strengthens the model's adaptability in complex scenes and remarkably enhances detection precision.

To thoroughly evaluate the model's performance in road crack detection before and after optimization, a comparative PR curve was generated with an IOU threshold of 0.5 during the testing phase. The results are presented in Figure 10, highlighting the model's effectiveness across both stages.

The area under the PR curve is the AP value. It measures the overall detection performance of the model by integrating or averaging the accuracy under different recall rates. In the context of object detection tasks, a larger AP indicates that the model has a lower false detection rate while accurately detecting objects, and thus exhibits better performance. Figure (b) clearly shows that the optimized model shows stronger detection ability.

### 3.4. Ablation Study

This study examines the performance enhancement of the YOLOv8-based road crack detection model through several innovative techniques. We introduced the BiFPN, the SimAM, and the SA-Net module to improve the model's accuracy and efficiency. The ablation study analyzed each component's contribution to the mAP, with results shown in Figure 9 and Table 2:

The baseline model, YOLOv8n, with an mAP of 41.5%, served as the control. After integrating the BiFPN module, the model's mAP increased to 42.6%, an enhancement of 1.1 percentage points. BiFPN combines feature information at different scales through its unique bidirectional feature fusion structure, improving the model's ability to detect small targets and optimizing feature flow, thereby reducing computational complexity and enhancing operational efficiency. Additionally, BiFPN's flexible weighted fusion among feature levels supports the complementary functionality of different features in object detection.

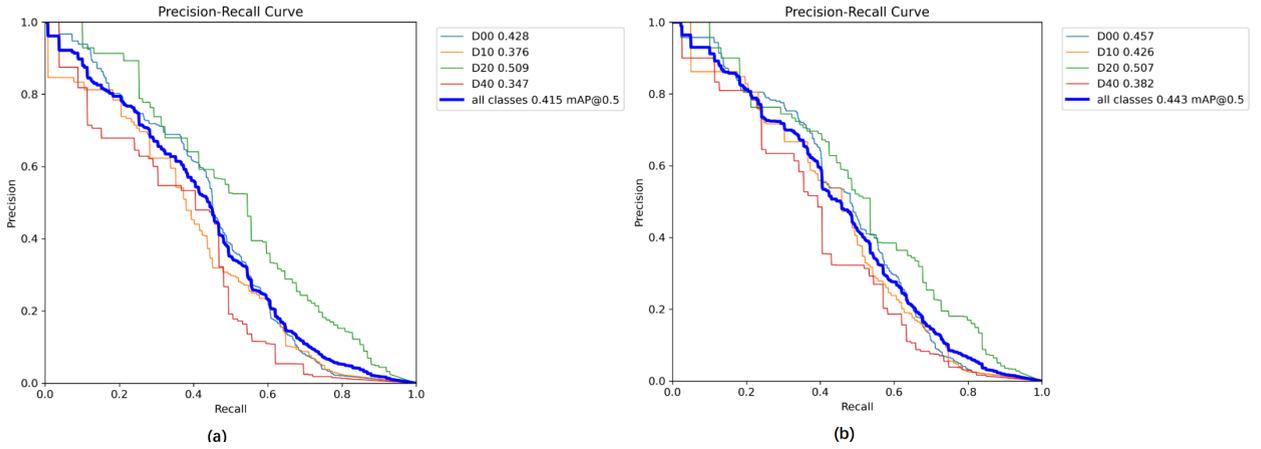

Figure 9: PR diagram of YOLOv8n and BsS-YOLO: **(a)**YOLOv8n; and **(b)**BsS-YOLO

Table 2
Results of the ablation experiments

| Model | Precision (%) | Recall (%) | F1 (%) | mAP (%) |
| --- | --- | --- | --- | --- |
| YOLOv8n | 52.1 | 44.3 | 47.9 | 41.5 |
| YOLOv8n+BiFPN | 54.2 | 45.6 | 49.5 | 42.6 |
| YOLOv8n+BiFPN+SimAM | 55.3 | 46.8 | 50.7 | 43.5 |
| YOLOv8n+BiFPN+SimAM+SA | 57.3 | 47.3 | 51.8 | 44.3 |

Introducing SimAM further raised the mAP to 43.5%, a gain of 0.9 percentage points. SimAM enables the model to concentrate on key features while disregarding irrelevant information through channel attention, significantly improving feature representation accuracy with minimal computational resources. Its lightweight design offers substantial accuracy improvements without increasing model complexity, making it ideal for resource-constrained applications.

Incorporating the SA-Net module led to a significant mAP increase to 44.3%, an improvement of 0.8 percentage points. SA-Net combines channel and spatial attention mechanisms, enhancing the model's discriminative ability by adaptively adjusting feature weights. This dual attention mechanism effectively captures crucial features in complex scenarios, improving generalization capability. Moreover, The lightweight design of SA-Net ensures minimal parameter overhead while enhancing accuracy, making it well-suited for diverse tasks including image classification, object detection, and image segmentation.

By incrementally introducing BiFPN, SimAM, and SA-Net, significant improvements in model performance were observed. These modules collectively enhanced crack detection accuracy while maintaining model efficiency and a lightweight structure, making the model suitable for real-time detection in practical applications. This research offers new technical pathways and optimization strategies for road crack detection, with potential applications in broader scenarios.

### 3.5. Mainstream Algorithm Experiments

To assess the performance of various models in specific tasks, we carried out a series of experiments, with results shown in Figure 10 and Table 3:

As illustrated in Table 3 and Figures 11 ,Unlike Faster-RCNN, which uses a two-stage detection approach, SSD employs a single-stage detection method, directly performing classification and regression on multi-scale feature maps, which improves detection speed. This single-stage structure enhances SSD's ability to capture targets of various sizes, resulting in an mAP of 47.6%. Compared to SSD, YOLOv3 improved multi-scale detection techniques and object box prediction methods, utilizing a deeper feature extraction network and adopting Darknet-53 as the backbone. These advancements increased YOLOv3's mAP to 48.2%, particularly excelling in complex scenarios.These models optimized the network structure by adopting more efficient architectures and advanced loss functions. YOLOv5 introduced adaptive anchors and new data augmentation techniques, while YOLOv7 focused on lightweight design while maintaining high accuracy. As a result, YOLOv5 and YOLOv7 achieved mAPs of 49.3% and 50.9%, respectively, demonstrating significant improvements over YOLOv3 in detection speed and accuracy. As a further optimized version of YOLOv8, BsS-YOLO incorporates advanced feature extraction networks and attention mechanisms to enhance road crack detection performance. These enhancements led to BsS-YOLO achieving the highest mAP of 57.3%, demonstrating its exceptional performance in terms of both detection accuracy.

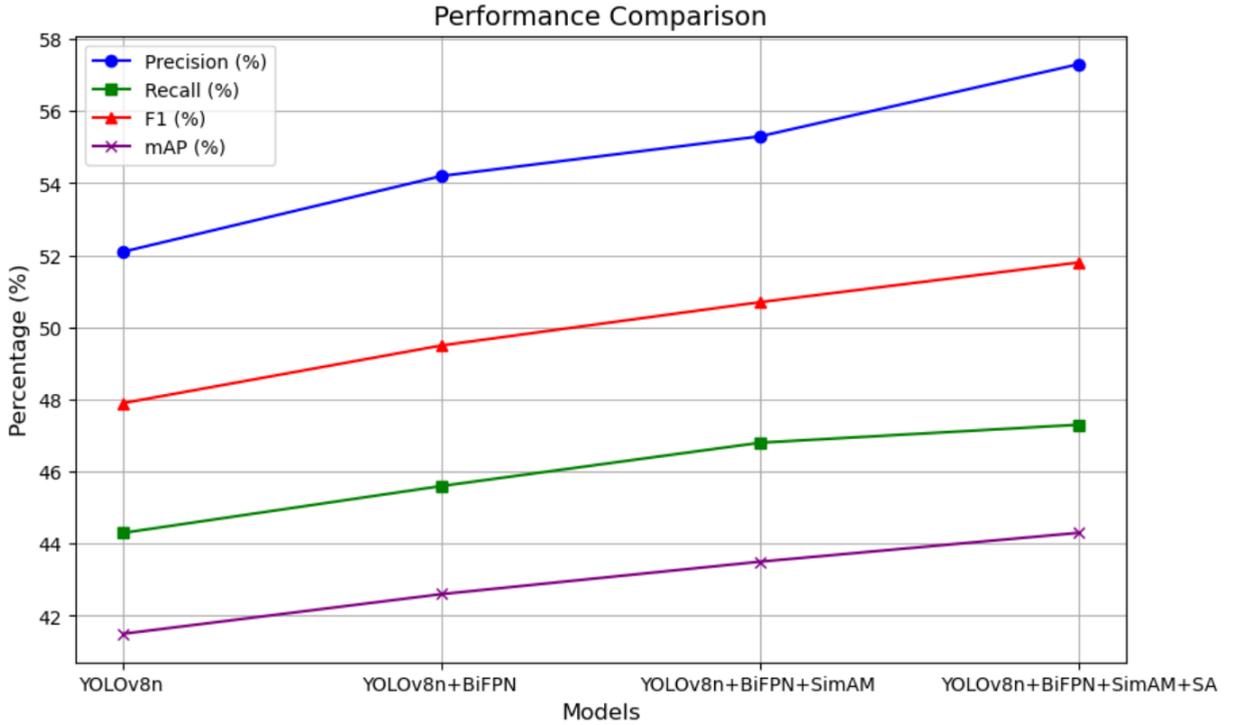

**Figure 10**: Comparative results of ablation experiments

Table 3
Comparison with Other Models

| Model | Precision (%) | Recall (%) | F1 (%) | mAP (%) |
| --- | --- | --- | --- | --- |
| Faster-RCNN | 44.5 | 37.5 | 40.7 | 35.9 |
| SSD | 47.6 | 38.6 | 42.6 | 36.5 |
| YOLOv3 | 48.2 | 40.2 | 43.8 | 38.2 |
| YOLOv5 | 49.3 | 41.5 | 45.1 | 39.1 |
| YOLOv7 | 50.9 | 42.3 | 46.2 | 40.6 |
| BsS-YOLO | 57.3 | 47.3 | 51.8 | 44.3 |

## 4. Discussion and Analysis

This paper introduces the BsS-YOLO model, developed for high-precision road crack detection. By integrating multi-scale feature fusion and attention mechanisms, this model effectively addresses challenges such as target scale variation and edge complexity, thereby enhancing detection accuracy and robustness.

First, the paper employs the BiFPN technique, which achieves multi-scale feature fusion through cross-scale connections and enhances feature representation via weighted feature fusion. Second, the SimAM attention mechanism is introduced, integrating spatial adaptive modules and channel attention mechanisms to optimize feature representation and enhance detection accuracy. Finally, Shuffle Attention is utilized to rearrange and mix features across channels, further strengthening the expression of key features. The collaborative effect of these innovative techniques results in a remarkable improvement in the mAP of the BsS-YOLO model for road crack detection.

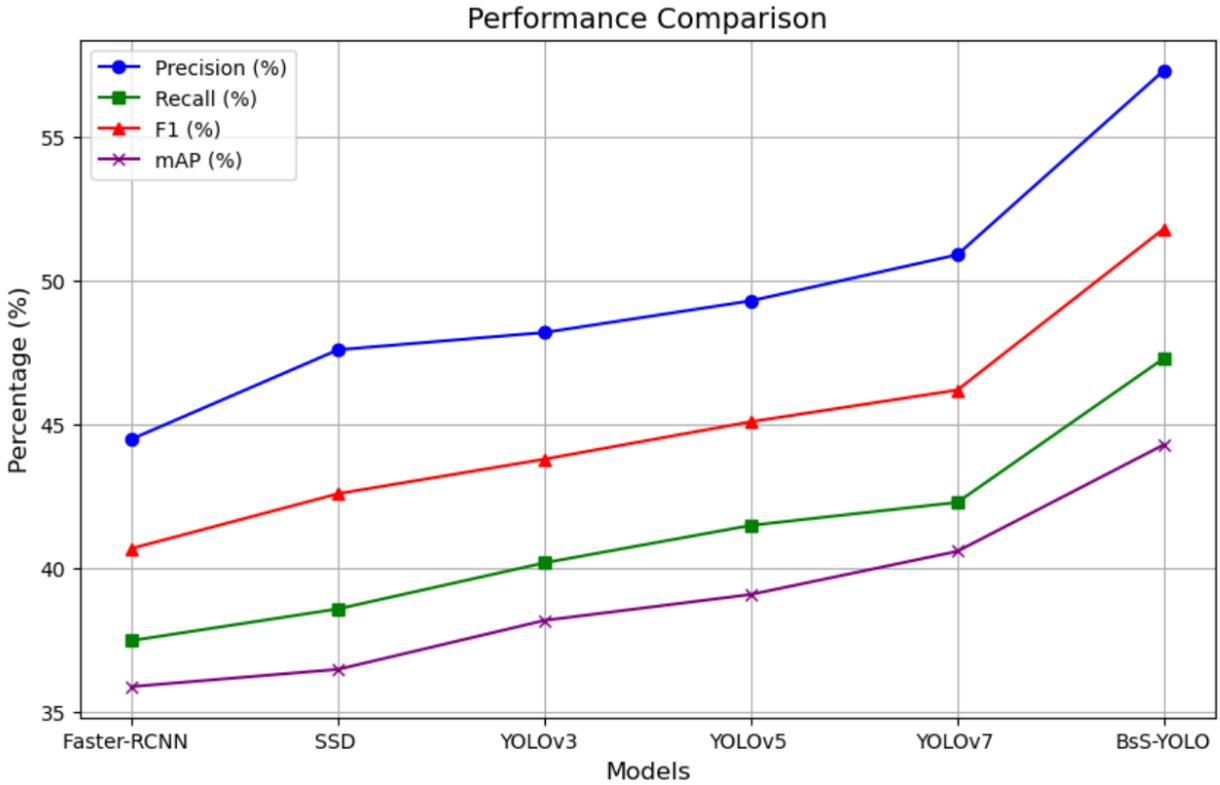

**Figure 11:** Comparison results with mainstream mode

However, the study still encounters several limitations. The dataset lacks diversity, particularly in crack images under complex weather and varying lighting conditions, which limits the model's generalization capability. Moreover, the annotation process for high-quality datasets is labor-intensive and prone to errors. While BiFPN improves detection capability and accuracy, it also increases computational load and memory consumption, limiting its suitability for real-time applications. The rapid weighted normalization fusion technique may struggle to consistently allocate optimal weights for feature maps of varying complexity, leading to unstable model performance. Additionally, the SimAM attention mechanism might not fully capture essential features in highly complex backgrounds, and Shuffle Attention could increase computational overhead when processing large-scale data.

Future research should focus on collecting data in complex environmental conditions and using data augmentation techniques to create new synthetic data, thereby expanding the training set and enhancing the model's generalization capability. Crowdsourcing and semi-automatic annotation tools could improve annotation quality. Furthermore, incorporating advanced techniques from other fields, such as Transformer models in deep learning, Graph Neural Networks (GNNs), Generative Adversarial Networks (GANs), and multi-task learning methods, could further optimize the model. Transformer models excel at handling long-range dependencies, self-supervised learning improves performance with limited labeled data, GNNs enhance irregular crack detection, GANs increase dataset diversity, and multi-task learning improves overall performance. These approaches will progressively enhance the model's detection accuracy, anti-interference ability and reliability , making it more effective in addressing the challenges of crack detection under complex road conditions.

## 5. Conclusion

Road crack detection is essential for maintaining urban infrastructure, yet current methods still require improvements in accuracy and efficiency. This paper introduces a model named BsS-YOLO, which integrates the BiFPN, SimAM attention mechanism, and Shuffle Attention. These innovations significantly enhance feature fusion efficiency and detection accuracy, leading to an 2.8% increase in mAP for road crack detection tasks. However, the model faces limitations, particularly in processing crack images under complex weather conditions and varying lighting, which constrain its generalization capability. While several new techniques have been incorporated, their robustness and adaptability in extreme conditions remain areas for further investigation. Future research should be centered on optimizing the model's performance in detecting small objects, expanding the dataset to encompass diverse weather conditions, and further improving attention mechanisms to enhance adaptability in dynamic environments.

## CRediT authorship contribution statement

**Jiaze Tang:** Conceptualization, Writing-Original draft, Software, Formal analysis, Project administration. **Angzehua Feng:** Data curation, Methodology, Investigation. **Vladimir Korkhov:** Supervision, Visualization. **Yuxi Pu:** Resources, Validation.